\documentclass[10pt,twocolumn,letterpaper]{article}

\usepackage{iccv}
\usepackage{times}
\usepackage{epsfig}
\usepackage{graphicx}
\usepackage{amsmath}
\usepackage{amssymb}
\usepackage{xurl}
\usepackage{booktabs}
\usepackage{multirow}
\usepackage{caption}
\usepackage[font=small,skip=0pt]{caption}
\usepackage{subcaption}
\usepackage{tablefootnote}
\usepackage{threeparttable}
\usepackage[numbers]{natbib}
\usepackage{hyperref}
\hypersetup{hidelinks,breaklinks=true,bookmarks=false}

\iccvfinalcopy 

\def\iccvPaperID{****} 

\ificcvfinal\fi

\begin{document}

\title{Fusing VHR Post-disaster Aerial Imagery and LiDAR Data for \\Roof Classification in the Caribbean}

\author{Isabelle Tingzon, Nuala Margaret Cowan, Pierre Chrzanowski\\
The World Bank Group, GFDRR\\
{\tt\small \{tisabelle, ncowan, pchrzanowski\}@worldbank.org}
}

\maketitle
\ificcvfinal\thispagestyle{empty}\fi

\begin{abstract}
Accurate and up-to-date information on building characteristics is essential for vulnerability assessment; however, the high costs and long timeframes associated with conducting traditional field surveys can be an obstacle to obtaining critical exposure datasets needed for disaster risk management. In this work, we leverage deep learning techniques for the automated classification of roof characteristics from very high-resolution orthophotos and airborne LiDAR data obtained in Dominica following Hurricane Maria in 2017. We demonstrate that the fusion of multimodal earth observation data performs better than using any single data source alone. Using our proposed methods, we achieve F1 scores of 0.93 and 0.92 for roof type and roof material classification, respectively. This work is intended to help governments produce more timely building information to improve resilience and disaster response in the Caribbean. 

\end{abstract}

\section{Introduction}

Natural hazards such as hurricanes, floods, landslides, and volcanic activity have been increasing in frequency and intensity in the Caribbean over recent years, raising great concern regarding the disaster risk of vulnerable communities and exposed populations \cite{munozotker2018building}. When Hurricane Maria ravaged Dominica in 2017, over 28,000 homes (90\% of the housing stock) were either damaged or destroyed, accumulating damages and losses estimated at USD 380M in the housing sector alone \cite{GoCD2017}. The destruction of structural assets as a result of extreme hazard events poses significant challenges to many small island developing states in the Caribbean. Extensive damage to buildings can lead to precarious housing conditions, building collapse, and loss of lives. Furthermore, the cost of repairing and rebuilding homes and public infrastructure can lead to severe budget reductions, increased debt, and ultimately, weaker economic growth prospects in the region \cite{GoCD2017}. 

Disaster risk management and mitigation planning are thus paramount to minimizing the adverse effects of natural hazards and preventing the loss of human lives \cite{gfdrr2023}. Disaster risk mitigation begins with analyzing the risk profile of buildings to different hazards and requires comprehensive information on the spatial distribution of buildings and their various characteristics. Building information modeling, as defined in \cite{wang2021machine}, involves the acquisition and management of the physical representation and characterization of built objects which form the basis for decision-making. Information on building characteristics is typically accessible via official government databases; however, due to the high costs and long timeframes associated with collecting and maintaining such datasets, up-to-date and granular building information is often lacking, inaccessible, or completely unavailable in many developing countries.

To address the challenge of data scarcity, researchers have turned to deep learning (DL) and earth observation (EO) for the automatic extraction of rich, fine-grained building attribute information. Recent studies have sought to classify buildings based on rooftop characteristics using convolutional neural networks (CNNs) in combination with different types of remote sensing data, such as very high-resolution (VHR) aerial imagery and light detection and ranging (LiDAR) data \cite{huang2022urban,buyukdemircioglu2021deep,partovi2017roof,olccer2023roof,castagno2018roof}. While the use of multimodal EO data presents new opportunities to extract more granular dimensions of building attribution, determining the appropriate data fusion strategy for integrating heterogenous image modalities remains a challenge \cite{dalla2015challenges}.

This study investigates the fusion of multimodal EO data for extracting rooftop characteristics in the Caribbean using DL techniques. Specifically, we explore two data fusion strategies: (1) feature-level data fusion and (2) decision-level integration using CNNs for classifying roof type and roof material from RGB orthophotos and LiDAR data in Dominica, taken after the advent of Hurricane Maria in 2017. We show that models trained using a combination of RGB and LiDAR images consistently outperform models trained using any one data source alone. Lastly, we evaluate the best models on an independent dataset of drone images with the goal of generating more frequently updated exposure datasets for disaster risk mitigation in the Caribbean.

\section{Application Context}

This work was developed in the context of the Digital Earth Project for Resilient Housing and Infrastructure in the Caribbean, a World Bank project funded by the Global Facility for Disaster Reduction and Recovery (GFDRR), which builds upon the work of the Global Program for Resilient Housing (GPRH) \cite{GPRH2019, WorldBank2022}. The project aims to enhance local capacity in the Caribbean to leverage EO-based solutions in support of resilient infrastructure and housing operations. This includes developing local skills and capabilities to produce and update critical building information needed for governments to improve resilience in the region.

The project also aims to support government initiatives such as the Resilient Housing Scheme by the Government of the Commonwealth of Dominica (GoCD), which strives to make 90\% of housing stock resilient by 2030, based on the Dominica Climate Resilience and Recovery Plan 2020-2030 \cite{government2020dominica}. For such programs to be successful, accurate and up-to-date maps of building characteristics are requisite for planning and monitoring the relocation of vulnerable citizens, the retrofitting of damaged structures, and the construction of new resilient homes \cite{government2020dominica}.

In line with these goals, the Digital Earth for a Resilient Caribbean Project is comprised of three components, described in further detail in Appendix \ref{appendix:a}: (1) capacity building, (2) generation and integration of baseline exposure datasets, and (3) knowledge exchange and dissemination. This paper primarily aims to advance the goals of the second component, which include leveraging DL and EO to fill baseline exposure data gaps in Caribbean countries. 


\section{Related Work} 
%


Recent years have seen a growing interest in leveraging DL and EO for vulnerability assessment of structural assets to better inform decision-making for disaster risk management \cite{linardos2022machine,silva2022building}. Previous studies demonstrate the successful applications of DL models, specifically CNNs, for characterizing buildings based on roof geometry (e.g. flat, hip, gable) from high-resolution satellite images and LiDAR data \cite{buyukdemircioglu2021deep,partovi2017roof,olccer2023roof,huang2022urban, castagno2018roof}. Recent studies have explored the use of CNNs for deep feature extraction in combination with ML algorithms for downstream roof type classification with promising results \cite{castagno2018roof,partovi2017roof}.

Likewise, CNNs have also been used in the context of roof material classification and damage assessment \cite{solovyev2020roof,kim2021cnn,naito2020building,xu2022damage,miura2020deep,kaur2023large,valentijn2020multi, kaur2023large}. Several studies have examined the applications of CNNs for classifying buildings into different roof material categories (concrete, metal, etc.) \cite{kim2021cnn,solovyev2020roof,norman2020fusion}. Other studies classify buildings based on the level of damage sustained, with recent works focusing on detecting the presence of blue tarpaulins from post-disaster satellite images \cite{xu2022damage,miura2020deep}. Our work seeks to expand this body of literature by evaluating CNN algorithms for rooftop classification using multimodal remote sensing data in the context of disaster risk management and post-disaster damage assessment in the Caribbean.

\section{Data}
To generate our ground truth dataset, we used the following three data sources for Dominica and Saint Lucia, as detailed in Table \ref{table:datasets}: (1) RGB aerial imagery, (2) LiDAR data, and (3) building footprints in the form of georeferenced vector polygons. RGB orthophotos and LiDAR data were obtained from the Government of Saint Lucia (GoSL) and GoCD; meanwhile, nationwide building footprints were delineated from aerial images by the World Bank Group. LiDAR-derived data products include the Digital Surface Model (DSM) and the Digital Terrain Model (DTM), from which we generate the normalized DSM (nDSM). Further details on data pre-processing can be found in Appendix \ref{appendix:lidar}.  

\begin{figure*}
\centering
\begin{subfigure}{.22\textwidth}
  \centering
  \includegraphics[width=0.85\linewidth]{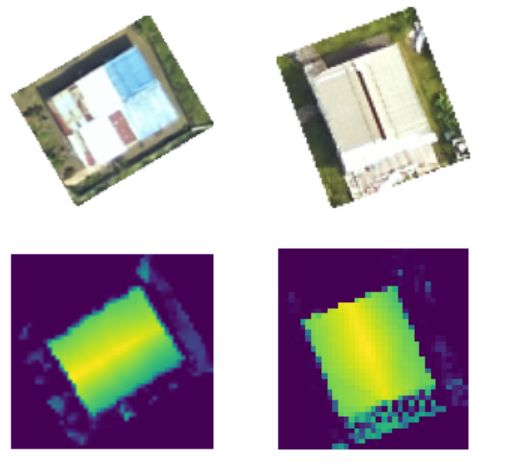}
  \caption{Gable}
  \label{fig:sub1}
\end{subfigure}%
\begin{subfigure}{.22\textwidth}
  \centering
  \includegraphics[width=0.83\linewidth]{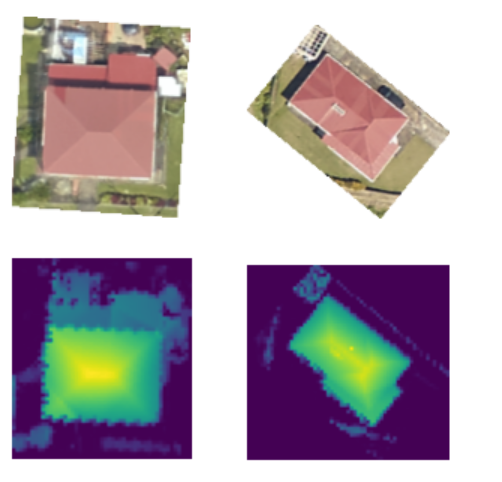}
  \caption{Hip}
  \label{fig:sub1}
\end{subfigure}%
\begin{subfigure}{.22\textwidth}
  \centering
  \includegraphics[width=0.7\linewidth]{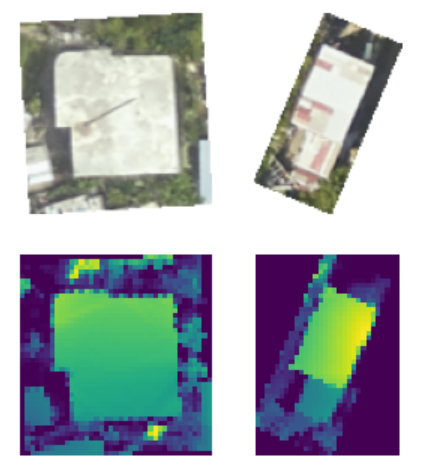}
  \caption{Flat}
  \label{fig:sub1}
\end{subfigure}%
\begin{subfigure}{.22\textwidth}
  \centering
  \includegraphics[width=0.85\linewidth]{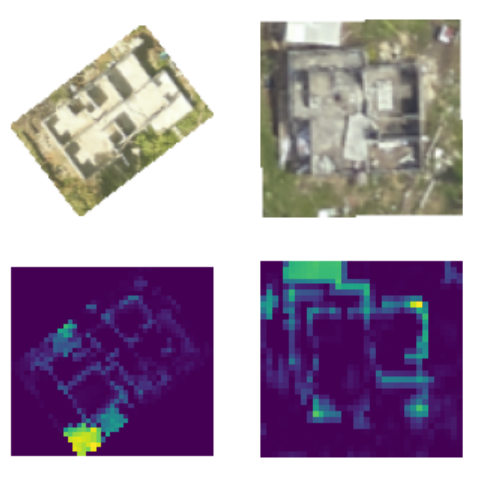}
  \caption{No roof}
  \label{fig:sub1}
\end{subfigure}%
\caption{Examples of RGB orthophotos (top) and LiDAR images (bottom) for different roof types.}
\label{fig:roof-type}
\end{figure*}

\begin{figure*}
\centering
\begin{subfigure}{.2\textwidth}
  \centering
  \includegraphics[width=1.0\linewidth]{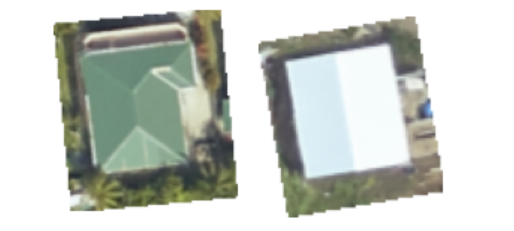}
  \caption{Healthy metal}
  \label{fig:sub1}
\end{subfigure}%
\begin{subfigure}{.2\textwidth}
  \centering
  \includegraphics[width=1.0\linewidth]{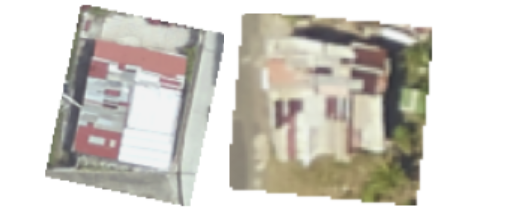}
  \caption{Irregular metal}
  \label{fig:sub1}
\end{subfigure}%
\begin{subfigure}{.2\textwidth}
  \centering
  \includegraphics[width=1.1\linewidth]{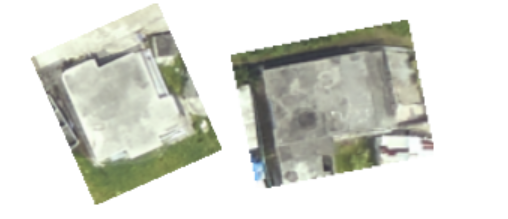}
  \caption{Concrete/cement}
  \label{fig:sub1}
\end{subfigure}%
\begin{subfigure}{.2\textwidth}
  \centering
  \includegraphics[width=1.1\linewidth]{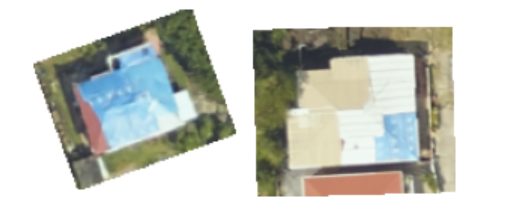}
  \caption{Blue tarpaulin}
  \label{fig:sub1}
\end{subfigure}%
\begin{subfigure}{.2\textwidth}
  \centering
  \includegraphics[width=1.0\linewidth]{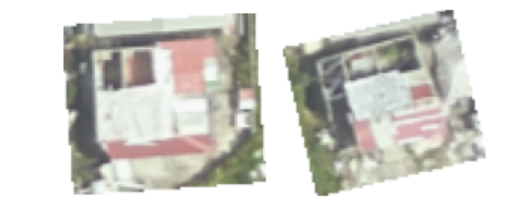}
  \caption{Incomplete}
  \label{fig:sub1}
\end{subfigure}%
\caption{Examples of RGB orthophotos for different roof material categories.}
\label{fig:roof-material}
\end{figure*}

To create a diverse and representative dataset, we selected 80 randomly sampled 500 m x 500 m tiles across Dominica and Saint Lucia and labeled each building within the selected tiles via visual interpretation of the RGB orthophotos and LiDAR images. We annotated a total of 8,345 buildings according to two attributes: (1) roof type and (2) roof material, the classes and distributions of which are listed in Table \ref{table:classdist}, based on the most common roof characteristics observed in the Caribbean \cite{openaicaribbean2019}. To generate the RGB and LiDAR image patches for each building, we crop the minimum bounding rectangle of the building footprint, scaled by a factor of 1.5, from each raster data source. Figures \ref{fig:roof-type} and \ref{fig:roof-material} illustrate examples of the RGB orthophotos and LiDAR-derived image patches. 


\begin{table}[t]
\caption{Details of the RGB orthophotos, LiDAR, and building footprints.}
\label{table:datasets}
\footnotesize
\resizebox{0.5\textwidth}{!}{%
\begin{tabular}{@{}lccccc@{}}
\toprule
 & \multicolumn{2}{c}{\textbf{RGB Orthophotos}} & \multicolumn{2}{c}{\textbf{LiDAR}} & \textbf{Buildings} \\ \midrule
 & \textbf{Resolution} & \textbf{Year} & \textbf{Spacing} & \textbf{Year} & \textbf{Count} \\
 \midrule 
\textbf{Dominica} & 0.20 m/px & 2018-2019  & 0.50 m & 2018-2019  & 50,347 \\
\textbf{Saint Lucia} & 0.10 m/px  & 2022 & 0.50 m & 2022  &  69,275 \\ \bottomrule
\end{tabular}}
\end{table}

\begin{table}[]
\footnotesize
\caption{Class distribution of roof type and roof material across countries and across the train/test splits.}
\label{table:classdist}
\begin{tabular}{@{}p{0.02cm}p{1.845cm}rrrrr@{}}
\toprule
                 & & \textbf{Dominica} & \textbf{Saint Lucia} & \textbf{Train} & \textbf{Test}  & \textbf{Total} \\ \midrule 
                 \multirow{4}*{\rotatebox{90}{\textbf{Roof Type}}}  
& \textbf{Gable}   & 2,055             & 1,063              & 2,339          & 779            & 3,118          \\
& \textbf{Hip}     & 1,334             & 710                & 1,534          & 510            & 2,044          \\
& \textbf{Flat}    & 1,636             & 274                & 1,433          & 477            & 1,910          \\
& \textbf{No Roof} & 1,145             & 128                & 955            & 318            & 1,273          \\
\midrule
\multirow{5}*{\rotatebox{90}{\textbf{Roof Material}}}  
& \textbf{Healthy metal}   & 1,420             & 1,343              & 2,073          & 690            & 2,763          \\
& \textbf{Irregular metal} & 1,206             & 507                & 1,285          & 428            & 1,713          \\
& \textbf{Concrete/cement} & 1,200             & 171                & 1,029          & 342            & 1,371          \\
& \textbf{Blue tarpaulin}  & 1,153             & 0                  & 865            & 288            & 1,153          \\
& \textbf{Incomplete}      & 1,191             & 154                & 1,009          & 336            & 1,345          \\
\midrule
& \textbf{Total}   & \textbf{6,170}    & \textbf{2,175}     & \textbf{6,261} & \textbf{2,084} & \textbf{8,345} \\ \bottomrule
\end{tabular}
\end{table}

\section{Methods}

\subsection{Baseline CNN Models}
We select ResNet50 \cite{he2016deep}, Inceptionv3 \cite{szegedy2016rethinking}, and EfficientNet-B0 \cite{tan2019efficientnet} as our CNN base architectures for experimentation \cite{castagno2018roof}. 
Each CNN is trained separately on RGB orthophotos and LiDAR-derived nDSM raster images; models trained using LiDAR are modified to accept single-channel image inputs. Further details on CNN model modifications and configurations, input image pre-processing, and data augmentation can be found in Appendix \ref{appendix:model-config}.

\subsection{Data Fusion}
\label{section:data-fusion}
We implement two fusion strategies for combining RGB orthophotos and LiDAR images (see Appendix \ref{appendix:data-fusion}) \cite{yin2021decision, dalla2015challenges}:
\begin{itemize}
    \item \textbf{Feature-level data fusion.} 
    Dense feature embeddings are extracted from the global average pooling layers of the best CNN model for each data source. The two feature vectors are then concatenated and used as input into a downstream ML classifier (see Figure \ref{fig:data-fusion}). 
    \item \textbf{Decision-level integration.} The results of the best-performing CNN per data source are combined using two different decision-level fusion strategies: (1) by computing the mean of the softmax vectors of the models, and (2) by concatenating the softmax layers from each CNN into a single vector and feeding this into a downstream ML classifier \cite{hoffmann2019model}. 
\end{itemize}
We experiment with the following ML classifiers for the downstream classification task: Logistic Regression (LR), Random Forests (RF), and Support Vector Machines (SVM). A detailed description of the  ML model hyperparameter tuning can be found in Appendix \ref{appendix:downstream-ml}.

\section{Results and Discussion}
\subsection{Model Evaluation}
For model training and evaluation, we split the dataset into 75\% training data and 25\% test data using stratified random sampling to preserve the percentage of samples for each class, as shown in Table \ref{table:classdist}. Note that the test data is comprised entirely of samples from Dominica, our primary region of interest in this study. To evaluate model performance, we report the precision, recall, F1 score, and accuracy (see Appendix \ref{appendix:model-eval} for more details). 

Table \ref{table:results1} shows the performance of CNNs trained individually on RGB orthophotos and LiDAR data for each classification task. Results indicate that for roof type classification, the LiDAR-only models perform slightly better than the RGB-only models; however, for roof material classification, the opposite is true, i.e. RGB-only models perform considerably better than LiDAR-only models with a difference of $>$25\% F1 score. 

We experimented with several data fusion techniques as shown in Table \ref{table:results2} and find that models trained using a fusion of RGB and LiDAR data yield a 1-3\% boost over models trained using any one data source alone. Furthermore, we find that feature-level data fusion consistently outperforms decision-level integration for both classification tasks. Our best model for roof type classification is an LR model that fuses the deep feature embeddings of a ResNet50 model trained on RGB and an Inceptionv3 model trained on LiDAR, achieving an F1 score of 93\%. Meanwhile, the best model for roof material classification is an RF model that fuses the feature embeddings of an EfficientNet-B0 model trained on RGB and an Inceptionv3 model trained on LiDAR data, achieving a 92\% F1 score.

\begin{table}[]
\small
\caption{Comparison of CNN test set results (\%) for (a) roof type classification and (b) roof material classification when trained individually on RGB and LiDAR image inputs.}
\label{table:results1}
\resizebox{0.475\textwidth}{!}{
\begin{tabular}{@{}p{0.02cm}lcccc@{}}
\multicolumn{4}{@{}l}{(a) Roof type classification}\\
\toprule
& & \textbf{F1 score} & \textbf{Precision} & \textbf{Recall} &  \textbf{Accuracy} \\ \midrule
\multirow{3}*{\rotatebox{90}{\textbf{\footnotesize{{RGB}}}}}  
& \textbf{ResNet50}    &   \textbf{89.04}                &  88.56                  &    89.61             &   88.72                \\
& \textbf{Inceptionv3}     &   89.02                &      89.01              &        89.04         &  88.53                 \\ 
& \textbf{EfficientNet-B0} &       87.34      &     87.12         &      87.63      & 87.04 \\
\midrule
\multirow{3}*{\rotatebox{90}{\textbf{\footnotesize{LiDAR}}}}  
& \textbf{ResNet50}      &       89.71            &    89.55                &        89.88         &    89.97               \\
& \textbf{Inceptionv3} &     \textbf{90.21}        &       90.09       &    90.37        &  90.60 \\ 
& \textbf{EfficientNet-B0} &     87.77        &     87.81         &   87.84         & 87.81 \\
\bottomrule
\end{tabular}}
\medskip \\
\resizebox{0.475\textwidth}{!}{
\begin{tabular}{@{}p{0.02cm}lcccc@{}}
\multicolumn{4}{@{}l}{(b) Roof material classification}\\
\toprule
&  & \textbf{F1 score} & \textbf{Precision} & \textbf{Recall}  & \textbf{Accuracy} \\ \midrule
\multirow{3}*{\rotatebox{90}{\textbf{\footnotesize{{RGB}}}}}  
& \textbf{ResNet50}     &    90.35               &             90.09       &        90.64         &   90.12                \\
& \textbf{Inceptionv3}   &       90.34            & 90.15                    &      90.70           &    90.07               \\ 
& \textbf{EfficientNet-B0} &    \textbf{90.69}         &   90.80           &   90.63         & 90.40 \\
\midrule
\multirow{3}*{\rotatebox{90}{\textbf{\footnotesize{LiDAR}}}}  
& \textbf{ResNet50}    &      62.22             & 61.32                   &       63.40          & 63.29                  \\
& \textbf{Inceptionv3} &   \textbf{64.01}          &  63.32            &   65.05         & 63.77 \\ 
& \textbf{EfficientNet-B0} &     61.28        &    60.49          &  62.55          & 61.13 \\
\bottomrule
\end{tabular}}
\end{table}

\begin{table}[]
\small
\caption{Test set results (\%) using (a) feature-level data fusion and (b) decision-level integration. For simplicity, we fuse the best-performing models for each task (see Table \ref{table:results1}).}
\label{table:results2}
\resizebox{0.475\textwidth}{!}{
\begin{threeparttable}
\begin{tabular}{@{}p{0.5cm}lcccc@{}}
\multicolumn{4}{@{}l}{(a) Feature-level data fusion}\\
\toprule
& & \textbf{F1 score} & \textbf{Precision} & \textbf{Recall} &  \textbf{Accuracy} \\ \midrule
\multirow{3}*{\rotatebox{90}{\textbf{\footnotesize{{\begin{tabular}[c]{@{}c@{}}Roof \\ Type\end{tabular}}}}}}  
& \textbf{LR}\textsuperscript{a}    &  \textbf{93.05}          &     93.00         &      93.11      &       92.90          \\
& \textbf{RF}\textsuperscript{a}    &   92.84          &   92.88           &    92.82        &     92.75            \\
& \textbf{SVM}\textsuperscript{a}     &    92.75         &     92.72         &   92.79         &       92.61           \\ \midrule
\multirow{2}*{\rotatebox{90}{\textbf{\footnotesize{{\begin{tabular}[c]{@{}c@{}}Roof \\ Material\end{tabular}}}}}}  
& \textbf{LR}\textsuperscript{b}     &   91.31          &  91.05            & 91.60            & 90.83             \\
& \textbf{RF}\textsuperscript{b}    &  \textbf{91.66}           &     91.48          &     91.85       &     91.17            \\
& \textbf{SVM}\textsuperscript{b} &   91.36          & 91.10              & 91.64           & 90.88
\\ \bottomrule
\end{tabular}
\end{threeparttable}}
\medskip \\
\resizebox{0.475\textwidth}{!}{
\begin{threeparttable}
\begin{tabular}{@{}p{0.5cm}lcccc@{}}
\multicolumn{4}{@{}l}{(b) Decision-level integration}\\
\toprule
&  & \textbf{F1 score} & \textbf{Precision} & \textbf{Recall}  & \textbf{Accuracy} \\ \midrule
\multirow{4}*{\rotatebox{90}{\textbf{\footnotesize{{\begin{tabular}[c]{@{}c@{}}Roof \\ Type\end{tabular}}}}}}  
& \textbf{Mean}\textsuperscript{a}     &    \textbf{92.92}         &        92.76      &    93.12        &   92.95             \\
& \textbf{LR}\textsuperscript{a}   &     92.25        &      91.92        &     92.63       &        92.13       \\ 
& \textbf{RF}\textsuperscript{a}    &   92.76          &        92.47      &    93.09        &        92.71         \\
& \textbf{SVM}\textsuperscript{a}    &  92.02           &     91.73         &   92.34         &       91.89        \\ 
\midrule
\multirow{4}*{\rotatebox{90}{\textbf{\footnotesize{{\begin{tabular}[c]{@{}c@{}}Roof \\ Material\end{tabular}}}}}}  
& \textbf{Mean}\textsuperscript{b}      & 91.05            &     91.04         &  91.09          &    90.55         \\
& \textbf{LR}\textsuperscript{b}    &   91.31          &     91.15         &    91.49        &   90.79       \\ 
& \textbf{RF}\textsuperscript{b}     &    91.22         & 90.98             &   91.51         &    90.55             \\
& \textbf{SVM}\textsuperscript{b}    &   \textbf{91.36}          &     91.23         &   91.49         &    90.83          
\\ \bottomrule
\end{tabular}
\begin{tablenotes}
\item[a] \scriptsize{Fuses ResNet50 trained on RGB and Inceptionv3 trained on LiDAR.}
\item[b] \scriptsize{Fuses EfficientNet-B0 trained on RGB and Inceptionv3 trained on LiDAR.}
\end{tablenotes}
\end{threeparttable}}
\end{table}

\subsection{Drone Imagery}
\begin{figure}
    \centering 
    \begin{subfigure}{.4\textwidth}
    \centering 
    \includegraphics[width=1.\linewidth]{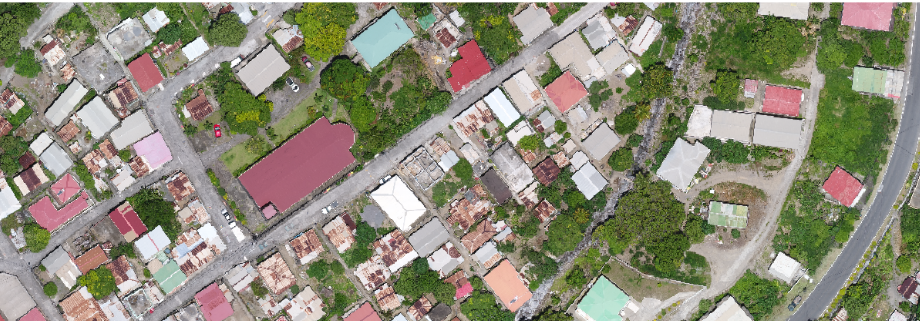}
    \caption{Pre-disaster drone image (2017).}
    \end{subfigure}
    \begin{subfigure}{.4\textwidth}
    \includegraphics[width=1.\linewidth]{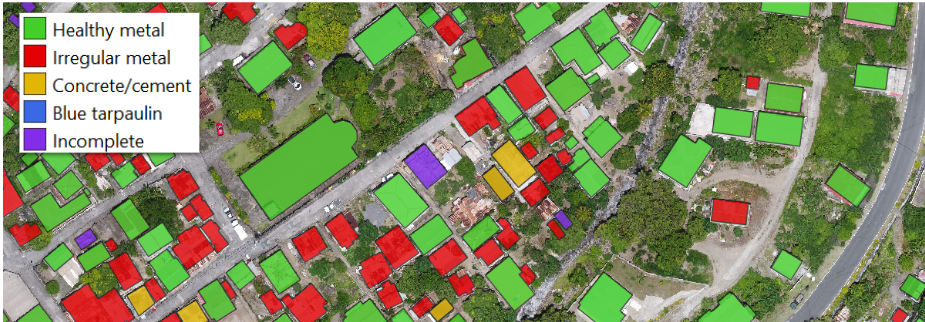}
    \caption{Pre-disaster roof material classification map.}
    \end{subfigure}
    \begin{subfigure}{.4\textwidth}
    \centering 
    \includegraphics[width=1.\linewidth]{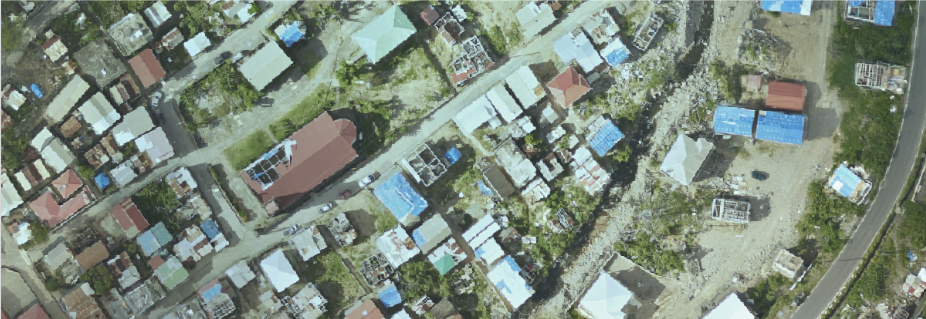}
    \caption{Post-disaster orthophoto (2018-2019).}
    \end{subfigure}
    \begin{subfigure}{.4\textwidth}
    \includegraphics[width=1.\linewidth]{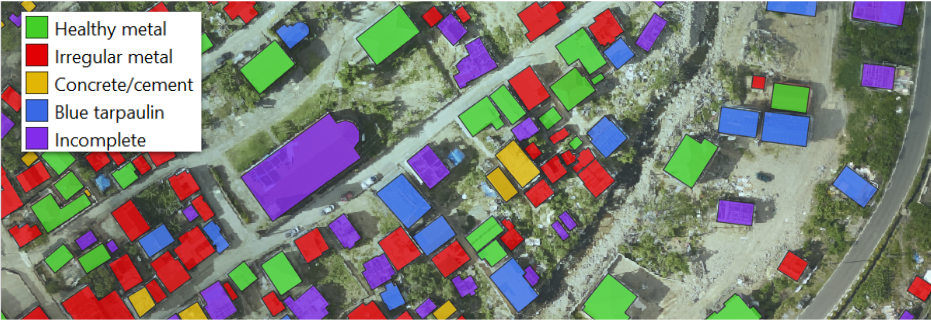}
    \caption{Post-disaster roof material classification map.}
    \end{subfigure}
    \caption{Pre- and post-disaster roof material classification maps of Colihaut, Dominica.}
    \label{fig:pre-post-disaster}
\end{figure}

One major component of the Digital Earth for a Resilient Caribbean Project involves training local communities to operate drones to enable frequent collection of VHR aerial images. To assess whether our models can generalize to VHR RGB drone imagery (3 cm/px), we manually labeled 373 buildings from pre-disaster drone images taken in Colihaut, Dominica in 2017. We evaluated our best RGB models using these annotations as ground truth and achieved F1 scores of 86\% for roof type classification and 90\% for roof material classification. Figure \ref{fig:pre-post-disaster} illustrates the pre- and post-disaster roof material classification maps in Colihaut. 

\section{Conclusion}
This study investigates different strategies for fusing RGB orthophotos and LiDAR images for roof type and roof material classification in the Caribbean. Our findings indicate that feature-level data fusion yields a 1-3\% boost over using any single data source alone. Our best models achieve F1 scores $>$ 90\% for both classification tasks. Future works will explore the use of additional data sources (e.g. drones, street-view images) for extracting building characteristics for climate resilience in the Caribbean.


\section*{Acknowledgments}
The Digital Earth Project for Resilient Infrastructure and Housing in the Caribbean is funded by the Global Facility for Disaster Reduction and Recovery (GFDRR). We are grateful for the support of the Government of Saint Lucia (GoSL) and the Department of Physical Development and Urban Renewal (DPDUR) in providing the datasets for Saint Lucia. We also thank the Government of the Commonwealth of Dominica (GoCD) for providing the data for Dominica and implementing component 2 of the Disaster Vulnerability Reduction Project (DVRP), funded by the World Bank Group and Climate Investments Fund (CIF) under the Pilot Program for Climate Resilience (PPCR). We are grateful for the insightful discussions with Sarah Antos as well as the initial work of the Global Program for Resilient Housing (GPRH), which provided the essential foundations for the study. We thank Michael Fedak and Christopher Williams for their assistance in providing access to the datasets as well as the insightful discussions on the data landscape in the Caribbean.

{\small
\bibliographystyle{ieee_fullname}

}

\appendix
\section*{Appendix}
\addcontentsline{toc}{section}{Appendices}
\renewcommand{\thesubsection}{\Alph{subsection}}

\subsection{Digital Earth for a Resilient Caribbean} \label{appendix:a}
The Digital Earth Project for Resilient Housing and Infrastructure in the Caribbean aims to support the development, scaling, and standardization of EO solutions for disaster risk management and is comprised of three main components:
\begin{enumerate}
    \item \textbf{Capacity building}. The project supports the training and upskilling of government officials, local community members, researchers, and other key stakeholders in geospatial data analytics, ML modeling, community mapping, and geospatial data management. 
    \item \textbf{Generation of critical exposure datasets}. The project strives to provide operational support for the generation, integration, and maintenance of new baseline exposure data layers through a combination of advanced technologies (e.g. AI and EO-based solutions) and local interventions (e.g. field surveys, participatory mapping). 
    \item \textbf{Knowledge exchange and dissemination}.  The project aims to promote the methodologies and results to other Caribbean regions through information exchange and knowledge-sharing sessions.
\end{enumerate}

\subsection{LiDAR Data Pre-processing}
\label{appendix:lidar}
The LiDAR-derived data products are provided in the form of the digital surface model (DSM) and digital terrain model (DTM). DSMs are a type of elevation model that represents the topography of the earth's surface, including artificial, man-made, or natural features such as tree tops, power lines, and buildings above bare ground. DTMs represent the bare-earth model absent of any natural or human-made features. The normalized DSM (nDSM) is calculated as the difference between the DSM and DTM and represents the relative height of features above the ground surface. nDSMs have been used in previous studies to improve the detection of buildings and building damage \cite{buyukdemircioglu2022deep,menderes2015automatic}. In our study, we use the single-channel nDSMs as input to CNNs for roof type and roof material classification. 

\subsection{CNN Model Configuration}
\label{appendix:model-config}
All models were pre-trained on the ImageNet dataset and fine-tuned using cross-entropy loss. For models trained using LiDAR-derived nDSM images, the initial convolutional layer was modified to accept single-channel input images, and its corresponding parameters were replaced with the mean of the weights from the pretrained model. Input images are zero-padded to a square up to the maximum between the width and height of the image and resized to 224 x 224 px for ResNet50 and EfficientNet-B0 and 299 x 299 px for Inceptionv3. We implement data augmentation in the form random vertical and horizontal image flips and rotations ranging from $-90^{\circ}$ to $90^{\circ}$.  For model configuration, we used the Adam optimizer and set the batch size to 32, with an initial learning rate of $1e^{-5}$. For the ResNet50 model, a dropout layer that sets input units to zero with a probability of 0.5 was added before the fully connected (FC) layer as a regularization mechanism. All models were trained for a maximum of 30 epochs, with a learning rate scheduler that reduces the learning rate by a factor of 0.1 after 7 epochs of no improvement.

\subsection{Data Fusion}
\label{appendix:data-fusion}
\begin{figure*}[h]
    \centering 
    \begin{subfigure}{.7\textwidth}
    \includegraphics[width=1.\linewidth]{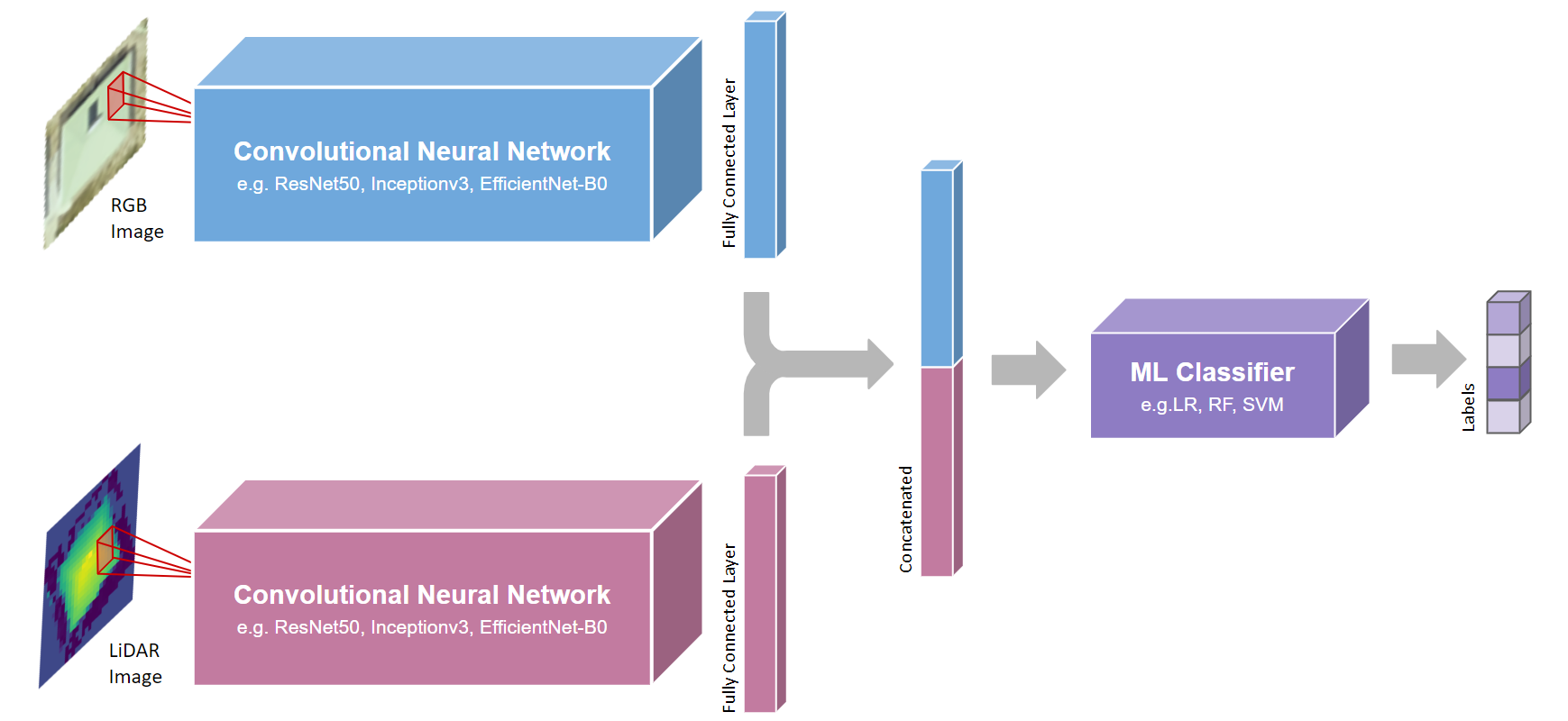}
    \caption{Feature-level data fusion}
    \end{subfigure}
    \begin{subfigure}{.587\textwidth}
    \includegraphics[width=1.\linewidth]{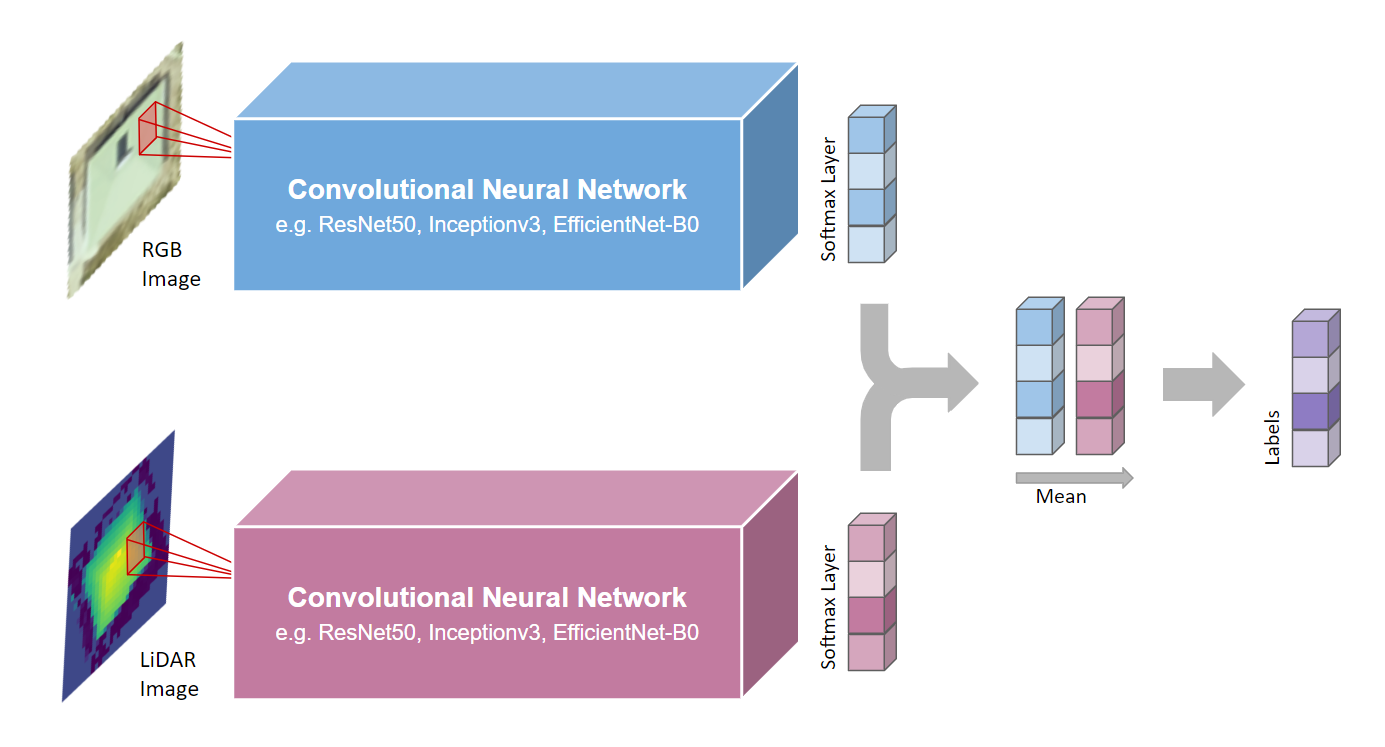}
    \caption{Decision-level data fusion (mean of softmax layers)}
    \end{subfigure}
    \begin{subfigure}{.7\textwidth}
    \centering 
    \includegraphics[width=1.\linewidth]{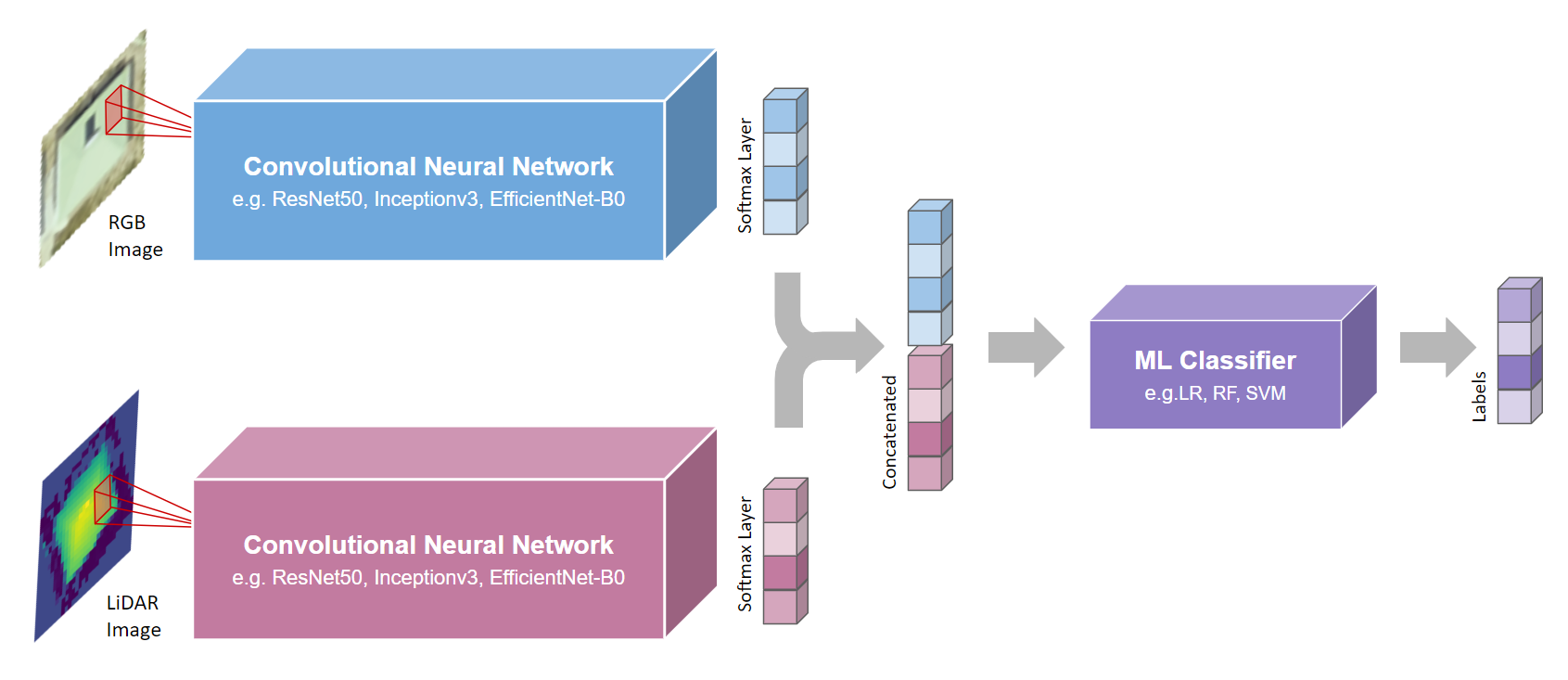}
    \caption{Decision-level data fusion (concatenated softmax layers)}
    \end{subfigure}
    \caption{Illustrations of feature-level data fusion (a) and decision-level data fusion (b and c). }
    \label{fig:data-fusion}
\end{figure*}

Based on the data fusion techniques described in Section \ref{section:data-fusion}, we extract two types of vector layers from the CNN models trained individually on each data source: (1) for feature-level data fusion, deep feature embeddings of size $(2048 \times 1)$ for ResNet50 and Inceptionv3 and $(1280 \times 1)$ for EfficientNet-B0 are extracted from the global average pooling layer of each CNN model and concatenated to form a single vector, and (2) for decision-level integration, a vector of probabilities is extracted from the softmax layer of each CNN and combined into a single vector. The combined vectors are then fed as input into an ML classifier for the downstream tasks of roof type and roof material classification.

For simplicity, we fuse the vectors (i.e. feature embeddings or softmax probabilities) extracted from the RGB and LiDAR models that achieve the best performance in terms of the F1 score for each of the classification tasks. Specifically, for roof type classification,  we fuse vectors extracted from the ResNet50 model trained on RGB images (F1 score: 89.04\%) and the  Inceptionv3 model trained on LiDAR images (F1 score: 90.21\%). Meanwhile, for roof material classification, we fuse the vectors of the EfficentNet-B0 model trained on RGB images (F1 score: 90.69\%) and the Inceptionv3 model trained on LiDAR images (F1 score: 64.01\%). 

\subsection{Downstream ML Classifiers}
\label{appendix:downstream-ml}
For the downstream classification task, we experiment with different ML methods, including LR, RF, and linear SVMs. We implemented hyperparameter tuning on the training set using a 5-fold cross-validation (CV) strategy, specifically: (a) grid search CV for LR and SVM and (2) randomized search CV for RF, with 30 parameter settings sampled set for each run. For LR and SVM, our search space includes the norm of the penalty (L1 and L2) and the regularization parameter $C$ (values include 0.001, 0.01, 0.1, and 1). For LR, we also experimented with two different types of solvers (LBFGS and LIBLINEAR). For RF, our search space includes the number of trees (set to values between 100 and 1000 with increments of 50), the maximum depth of trees (set to values between 3 and 10), and the criterion measuring the quality of the split (gini or entropy). For all models, we tried different scaling techniques, including min-max scaling, standard scaling, and robust scaling as implemented in \textit{scikit-learn}. 

For roof type classification, the best F1 score of 93\% is achieved by an LR model with an L2 penalty term and a 0.1 regularization parameter $C$. The model fuses the concatenated feature embeddings (min-max scaled) of two models: a ResNet50 model trained on RGB and an Inceptionv3 model trained on LiDAR. For roof material classification, the best F1 score of 92\% is achieved by an RF model with 450 estimators, a maximum depth of 9, and an entropy criterion. The RF model is trained on the combined feature embeddings (min-max scaled) of an EfficientNet-B0 model trained on RGB and an Inceptionv3 model trained on LiDAR data.

\subsection{Performance Metrics} 
\label{appendix:model-eval}
We compute precision, recall, F1 score, and accuracy using standard definitions as follows:
\begin{equation}
    \text{Precision} = \frac{\text{TP}}{\text{TP} + \text{FP}}
\end{equation}
\begin{equation}
    \text{Recall} = \frac{\text{TP}}{\text{TP} + \text{FN}}
\end{equation}
\begin{equation}
    \text{F1 score} = \frac{2 \cdot \text{Precision} \cdot \text{Recall}}{\text{Precision} + \text{Recall}}
\end{equation}
\begin{equation}
    \text{Accuracy} = \frac{\text{TP} + \text{TN}}{\text{TP} + \text{TN} + \text{FP} + \text{FN}}
\end{equation}
where $\text{TP}$ is the number of true positives, $\text{TN}$ is the number of true negatives, $\text{FP}$ the number of false positives, and $\text{FN}$ the number of false negatives. The precision, recall, and F1 score metrics are computed as the unweighted mean of the values calculated per class (i.e. macro-averaged). 

\end{document}